\documentclass[letterpaper, 10 pt, conference]{ieeeconf}  

\IEEEoverridecommandlockouts                              

\usepackage{amsmath} 
\usepackage{amssymb}  
\usepackage{makecell}
\usepackage{hyperref}
\usepackage{tabularx}
\usepackage{cite}
\usepackage{graphicx}
\usepackage{booktabs}
\usepackage{caption}

\title{\LARGE \bf
Learning Adaptive Hydrodynamic Models Using Neural ODEs in Complex Conditions
}

\author{Cong Wang$^{1*}$, Aoming Liang$^{2*}$, Fei Han$^{2}$, Xinyu Zeng$^{2}$, Zhibin Li$^{3}$, Dixia Fan$^{2}$, and Jens Kober$^{1}$
\thanks{*These authors contributed equally}
\thanks{$^{1}$Authors are with the Department of Mechanical Engineering, TU Delft, Delft, The Netherlands
        }%
\thanks{$^{2}$Authors are with the School of Engineering, Westlake University, Hangzhou, Zhejiang, China
        }%
\thanks{$^{3}$Author is with the Department of Computer Science, University College London, London, United Kingdom
        }%
}

\begin{document}

\maketitle
\thispagestyle{empty}
\pagestyle{empty}

\begin{abstract}

Reinforcement learning-based quadruped robots excel across various terrains but still lack the ability to swim in water due to the complex underwater environment.
This paper presents the development and evaluation of a data-driven hydrodynamic model for amphibious quadruped robots, aiming to enhance their adaptive capabilities in complex and dynamic underwater environments. 
The proposed model leverages Neural Ordinary Differential Equations (ODEs) combined with attention mechanisms to accurately process and interpret real-time sensor data.
 The model enables the quadruped robots to understand and predict complex environmental patterns, facilitating robust decision-making strategies. We harness real-time sensor data, capturing various environmental and internal state parameters to train and evaluate our model. A significant focus of our evaluation involves testing the quadruped robot's performance across different hydrodynamic conditions and assessing its capabilities at varying speeds and fluid dynamic conditions. The outcomes suggest that the model can effectively learn and adapt to varying conditions, enabling the prediction of force states and enhancing autonomous robotic behaviors in various practical scenarios. 

\end{abstract}

\section{Introduction}
\label{sec:intro}


Quadruped robots have gained significant attention in recent years due to their potential applications in various fields, such as underground inspections, scientific exploration of challenging planetary analog environments, robust walking in the wild, and jumping and landing from air \cite{arm2023scientific, tranzatto2022cerberus, miki2022learning, DBLP:journals/ras/YaoWWMYZW23, DBLP:conf/robio/YaoWYWZZW22, rudin2021cat}. However, none of these robots currently possess the ability to operate in water. Accurate hydrodynamic modeling is essential for the operation of quadruped robots in water. 
This process is challenging due to complex fluid-structure interactions (FSI), the impact of varying leg configurations on fluid resistance, and the difficulty of accurately simulating the dynamic underwater environments.

Traditional hydrodynamic models, which rely on the numerical solution of the Navier-Stokes equations, are effective but come with high computational costs and may not efficiently handle complex boundary conditions or nonlinear dynamics \cite{tezduyar2006space, hirt1974arbitrary, mittal2005immersed}. These limitations hinder their practical application, especially for large-scale or real-time simulations \cite{newman2018marine}.
Fluid-structure interaction phenomena, found in many natural phenomena, such as insect wings and fish fins, are typically modeled using body-fitted grid \cite{hirt1974arbitrary}, space-time finite-element method \cite{tezduyar2006space} and immersed-boundary method \cite{mittal2005immersed}. However, these methods require extensive computational resources, limiting their application in practical problems. For robots with real-time changing structures, such as \cite{DBLP:journals/ral/YaoMZZPWLW23, thrust_allocation}, it is challenging to use fixed hydrodynamic coefficients for calculations. The fluid-structure interaction during motion is difficult to model accurately, which poses significant challenges for precise hydrodynamic modeling. These complexities necessitate using data-driven methods to predict the hydrodynamic forces and interactions in such dynamic environments.


\begin{figure}[t]
    \centering
    \includegraphics[width=\linewidth]{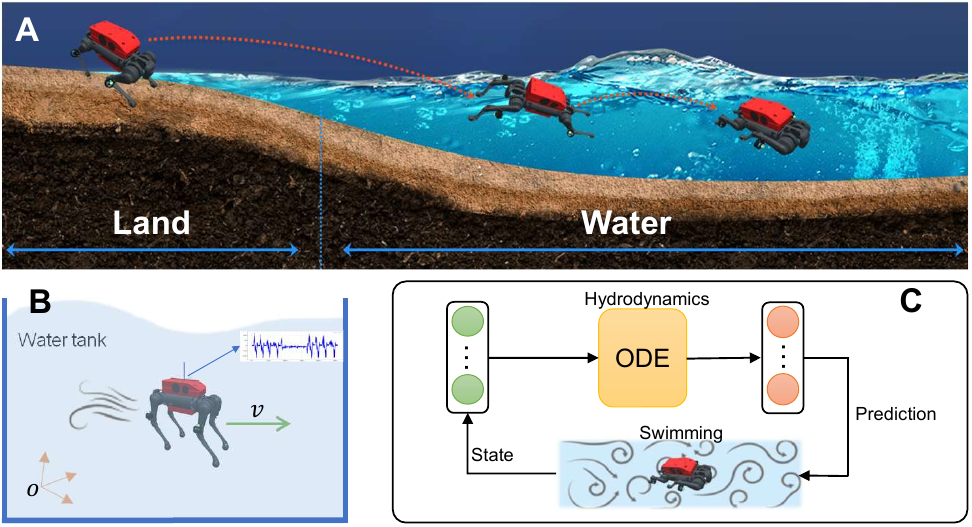}
    \vspace{-0.5cm}
    \caption{Overview. A: The amphibious SwimmingDog robot from land to water. B: Collect data in water tank. C: ODE-based hydrodynamic model.}
    \label{fig:overview}
    \vspace{-0.5cm}
\end{figure}

Recent advances in machine learning, particularly deep learning, offer promising alternatives to these traditional methods \cite{vinuesa2023transformative, brunton2020machine, eivazi2024physics, fan2019robotic, wang2021physics}. 
Neural Ordinary Differential Equations (ODEs) \cite{chen2018neural} represent a novel class of deep learning models that frame the dynamics in the form of differential equations, providing a natural and flexible approach to modeling time-continuous systems. Unlike other dynamic models like symplectic neural networks \cite{wen2022learning} and Lagrange neural networks \cite{cranmer2020lagrangian}, Neural ODEs show a unique advantage by parameterizing the derivative of the state with respect to time using neural networks. This allows for an adaptive computation of the dynamics \cite{liang2021modeling, kohler2019equivariant,lee2021parameterized,zakwan2023physically, jiahao2022learning,mo2024pi}, which could be particularly beneficial for capturing the intricate and nonlinear behaviors observed in the hydrodynamic environment. \cite{kidger2020neural} proposed a Controlled Neural ODEs(CNODE) to handle the irregular time series data. The ODE-Transformer, introduced by \cite{li2022ode}, merges ODE-based continuous modeling with the Transformer's discrete processing capabilities in machine translation tasks.

Integrating data-driven approaches with Neural ODEs enables the leveraging extensive experimental and simulated datasets, enhancing the capability to generalize and predict under varied conditions without the need for explicit formulation of the governing physical laws. This is crucial for scenarios where the physics is poorly understood in traditional equations \cite{zakwan2024neural,di2023simba,meleshkova2021application}. \cite{zhu2024data} propose the Auto-tuning Blimp-oriented Neural Ordinary Differential Equation method to tackle aerodynamic modeling in the miniature robotic blimps. \cite{ingebrand2024zero} leverage the theory of function encoders to rapidly identify dynamics in the learned space, which includes a set of basis functions in Neural ODEs. \cite{salehi2023adaptive} presented a meta-learning control method based on Neural ODEs for adaptive dynamics prediction in asynchronous industrial robots.



Machine learning-based approaches in fluid dynamics have demonstrated tremendous potential. Many researchers have focused on the dynamic modeling of robotic fish and bluff bodies \cite{youssef2022design,rabault2020deep}. \cite{colabrese2017flow} first demonstrated that reinforcement learning (RL) efficiently addresses Zermelo’s Problem. They adopted this method to train a point-like swimmer in an Arnold-Beltrami-Childress (ABC) flow to navigate vertically as quickly as possible. \cite{ren2021bluff} propose a novel active-flow-control strategy for bluff bodies to hide their hydrodynamic traces.  \cite{zhu2022learning} adopted multi-agent RL methods to learn a schooling behavior in two fishes. Although these methods yield satisfactory results in numerical computations, they are seldom used in practice. The primary reason is that it is hard to learn useful dynamic relationships, leading to poor performance during transitions between switching conditions. Transformers are a type of deep learning model that revolutionized the field of natural language processing \cite{vaswani2017attention} and have since been adapted to various other domains, including image processing \cite{parmar2018image}, robots \cite{shridhar2023perceiver,yu2023novel,chen2023transformer}, and time-series analysis \cite{wen2022transformers}. 
Attention not only facilitates effective information fusion but also enables reduced transformation. It significantly enhances the ability to understand and fuse information. \cite{khoshsirat2023transformer} design
a proven design element from top-performing networks, integrating transformer blocks as core building blocks in Neural ODEs. \cite{serifi2023transformer} proposes to augment simulation representations with a transformer-inspired architecture by training a network to predict the true state of robot building blocks given the simulation state in the robot application. Although the methods above have succeeded in deep learning fields, the transformer-based ODE model for underwater robots has not yet been studied.

To address these challenges, it is crucial to evaluate the model's performance under increasingly complex conditions that closely resemble real-world scenarios.
In this study, we design a series of tasks with escalating complexity to systematically evaluate our proposed data-driven hydrodynamic model.
This paper presents a practical approach to modeling the hydrodynamics of quadruped robots with swimming capabilities using Neural ODEs. To leverage the parallelization benefits of attention modules, we employ self-attention to compress the input bottleneck of the condition vector.
Throughout our subsequent simulation experiments, the attention module processes an input of up to high-dimensions. This study chooses self-attention to extract the feature in the condition vector. We make several key contributions: 

\textbf{1) Unique Dataset Collection.} We collect a comprehensive dataset through controlled towing experiments, providing valuable hydrodynamic data for quadruped robots. This dataset captures a wide range of motion scenarios, offering a robust foundation for training and validating hydrodynamic models.

\textbf{2) Attention-Based Neural ODE Framework.} We develop a Neural ODE framework integrated with attention mechanisms. This design was chosen to effectively capture the temporal dependencies and complex interactions between the robot's dynamic states and the resulting hydrodynamic forces, enhancing prediction accuracy. 

\textbf{3) Robust Prediction under Varying Conditions.} We demonstrate the model's ability to predict force states under varying conditions, including different speeds and quadruped leg configurations, showcasing its robustness and adaptability in dynamic scenarios. The attention mechanism allows the model to adapt to dynamic scenarios by focusing on relevant features, ensuring high adaptability and robustness in real-time applications.

To the best of our knowledge, this pioneering study provides significant advancements in fluid dynamics modeling for quadruped robots, particularly in underwater applications.

\section{Method}
\label{sec:method}

\begin{figure*}[t]
    \centering
    \includegraphics[width=1.0\linewidth]{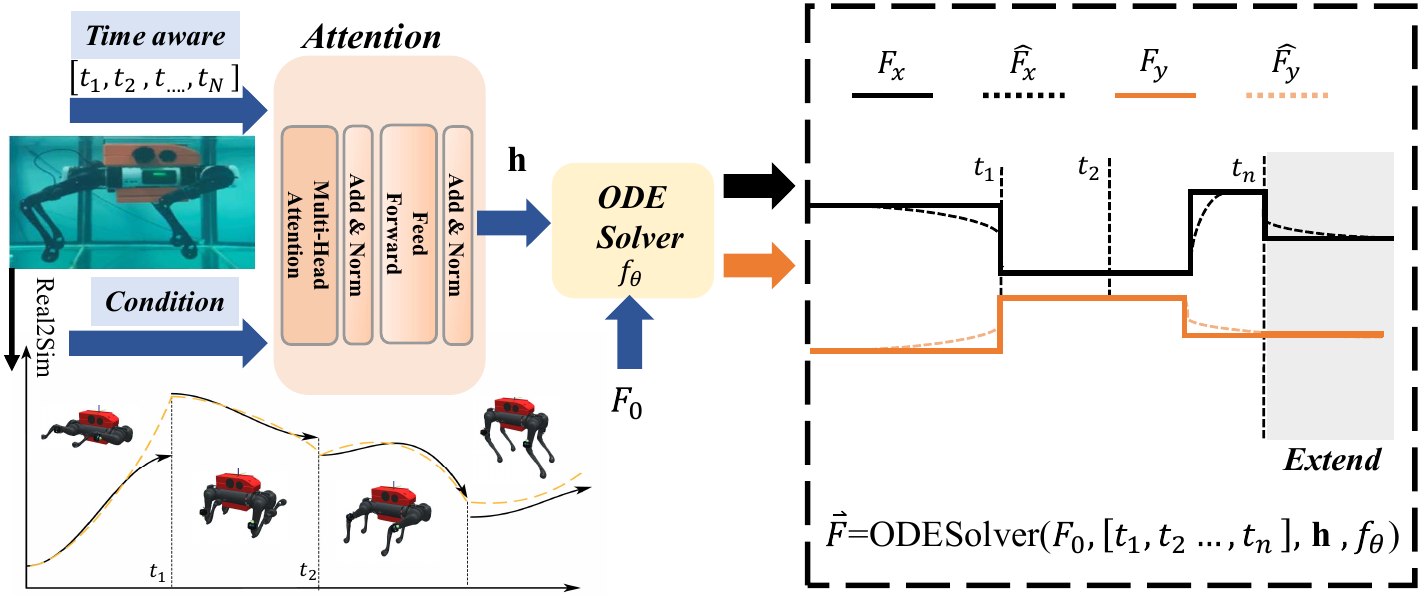}
    \caption{Method overview. Left: robot motion with different conditions in the time aware stamps [$t_1$,$t_2$,...,$t_N$]. Middle: the attention-based neural ODE extracts the kinematic information to latent vector
    $\mathbf{h}$. Right: the prediction of hydrodynamic force trajectory $\widehat{\mathbf{F}}$ is integrated by a ODESolver among [$t_1$,$t_2$,...,$t_L$] based on the the initial condition $\mathbf{F}_0$. The $f_{\theta}$ is a neural network. }
    \label{fig:method}
    \vspace{-0.5cm}
\end{figure*}

\subsection{Preliminary}

\paragraph{Neural ODEs based model}
Neural ODEs \cite{chen2018neural} learn the dynamics by learning a continuous transformation of the input space.  Neural ODEs can potentially capture the continuous-time changes in forces as a function of sensor inputs.

In deploying a Neural ODEs model, the approach differs from traditional discrete-time step models by treating the entire time series as a continuous flow of inputs into the model. A key feature of Neural ODEs is use of a parameterized ordinary differential equation \( \frac{d\mathbf{x}(t)}{dt} = f(\mathbf{x}(t),t, \theta) \) to model the dynamic behavior, where \( \mathbf{x}(t) \) is the vector of states and \( f \) is a function defined by the neural network with parameters \( \theta \).

To compute the state \( \mathbf{x}(t) \) at any time \( t \), from the given initial state \( \mathbf{x}(t_0) \),  it can be achieved by numerical integration methods such as the Forward Euler method and the Runge-Kutta method by ODE solvers \cite{butcher1996history}. 
\vspace{-0.25cm}
\begin{align}
\mathbf{x}(t) &= \mathbf{x}(t_0) + \int_{t_0}^{t} f(\mathbf{x}(\tau), \theta) d\tau \\
&= \operatorname{ODEsolver}(\mathbf{x}(t_0),f,t_0,t,\theta)
\end{align}

Where \( \tau \) represents the time variable. The choice of numerical integrator and step size can significantly affect the accuracy and stability of the model predictions. 

Consider minimizing a scalar-valued loss function $L()$. The Neural ODEs defines a adjoint state $\mathbf{a(t)} =  \frac{\partial L}{\partial \mathbf{x_{t}}} $, and its derivation could be defined as $\frac{d \mathbf{a}(t)}{d t}=-\mathbf{a}(t)^{\top} \frac{\partial f(\mathbf{x}(t), t, \theta)}{\partial \mathbf{x}}$. The Neural Network $f$ can use backpropagation to update its parameters. For more detail, please refer to \cite{chen2018neural}.

\paragraph{Transformer}
The core of the transformer \cite{vaswani2017attention} is the attention mechanism. In practical use, self-attention indicates that Query, Key, and Value are the same vector.  

\subsection{Model architecture}

Our proposed model architecture aims to predict the hydrodynamic forces acting on a quadruped robot using a sequence of observation data. This work seeks to develop a generalized machine-learning model adaptable to a wide range of underwater robot applications, ensuring versatility and broad applicability across different environments and conditions.
The key components of our model include Neural Ordinary Differential Equations (ODEs) and attention mechanisms. The steps involved in the model architecture are as follows:

\textbf{Input Data}: The input data consists of a sequence of observation kinematic data \(\mathbf{x}_1, \mathbf{x}_2, \ldots, \mathbf{x}_N \in \mathbb{R}^{n} \), each representing motion parameters at uniform time steps. In this work, the input dimension \(n\) can take values of either 4 or 35. When \(n = 4\), it corresponds to the actual measurements of two joint angles and the linear speeds along two axes. When \(n = 35\), it corresponds to data generated from the MuJoCo \cite{DBLP:conf/iros/TodorovET12} simulator in the real-to-sim setup.
    
\textbf{Attention Mechanism}: We incorporate an attention mechanism to effectively capture the temporal dependencies and dynamics in the observation data. The attention mechanism processes the input data into a latent representation. This transformation uses n-dimensional inputs and maps them to a latent vector \(\mathbf{h}\) \(\in \mathbb{R}^{h}\) dimensions. The attention layer enables the model to focus on hidden features, enhancing the learning of the underlying dynamics.  In the attention-based model, the hidden state is calculated as $\mathbf{h}$ = Attention$(\mathbf{x})$. If there is no attention, the $\mathbf{h}$ = MLP($\mathbf{x}$). Where the MLP is a feed-forward neural network.
    
\textbf{Neural ODE Framework}: The core of our model is the Neural ODE framework, which learns the kernel function \( f \) that represents the hydrodynamic dynamics. The Neural ODE is formulated as follows:
$$
\mathbf{\hat{F}}(t) = \mathbf{{F}}(t_0) + \int_{t_0}^{t} f(\mathbf{h}, \theta,\tau) d\tau
$$
    where \(\mathbf{\hat{F}}(t)\) is the predicted force vector, \( \mathbf{h} \) is the latent representation obtained from the attention layer, and \(\theta\) represents the parameters of the neural network.
    
\textbf{ODE Solver}: To compute the state \(\mathbf{F}(t)\) at any time \( t \), starting from the initial state \(\mathbf{F}(t_0)\), we use numerical integration methods such as the Forward Euler method and the Runge-Kutta method. The ODE solver integrates the kernel function \( f \) over time, generating a continuous flow of outputs.
\vspace{-0.25cm}
\begin{align}
    \mathbf{F}(t) &= \mathbf{F}(t_0) + \int_{t_0}^{t} f(\mathbf{F}(\tau), \theta) d\tau \\
    &= \operatorname{ODEsolver}(\mathbf{F}(t_0), f, t_0, t, \mathbf{h}, \theta)
\end{align}

\textbf{Prediction}: The output of the ODE solver is a sequence of predicted force vectors \( \mathbf{F}_1, \mathbf{F}_2, \ldots, \mathbf{F}_L \in \mathbb{R}^{f} \), where \( L \) is the prediction length that can be tailored to span different temporal scopes. \(f\) represents the dimensionality of the output. When \(f = 2\), the output corresponds to the forces along the \(x\) and \(y\) axes. When \(f = 6\), the output includes both the forces and moments from the simulation data. 

The adoption of an attention-based architecture for learning latent representations is both reasonable and innovative. Past Neural ODEs require the input state as the same as the output state like $\mathbf{F}$. However, the input consists of four-dimensional kinematic trajectories. During training, the attention layer maps postures and forces effectively. 
This implementation employs attention mechanisms to enhance the feature representation of kinematic data.

From Fig.~\ref{fig:overview}, the model incorporates the attention mechanism into model because the dimensionalities of sequence $\in \mathbb{R}^{n}$ and outputs $\in \mathbb{R}^{f}$  do not inherently conform to the rules of  ODEs. We perform a transformation that uses inputs with a latent representation $\mathbf{h} \in \mathbb{R}^{f}$
by mapping through the Attention layer, the Neural ODEs can effectively learn the latent vector derivation. 
In this work, we include two tasks. The first task has an input dimension of 4 and an output dimension of 2, hereafter referred to as \textit{Task 1}. The second task has an input dimension of 35 and an output dimension of 6, hereafter referred to as \textit{Task 2}. For \textit{Task 2}, we focus on long-term prediction over 400 time steps, with a 10\% perturbation for comparison. To facilitate differentiation, we will refer to these tasks as \textit{Task 1} and \textit{Task 2} in the following sections. Note that \textit{Task 1} consists of three different experiments.  The ODE solver begins with this true initial value $F(t_{0})$. 




\subsection{Learning Objective}

The learning objective is to minimize the prediction error between the predicted force vectors and the ground truth measurements. The steps involved in the learning process are as follows:

\textbf{Dataset Preparation}: The dataset consists of sequences of observation data and corresponding force measurements. The dataset is divided into training, validation, and test sets.
$$
\mathcal{D}:=\left\{\left(\hat{F}_0^j, {\mathbf{x}_0^j}\right), \left(\hat{F}_1^j, \mathbf{x}_1^j\right), \cdots, \left(\hat{F}_L^j, \mathbf{x}_L^j\right)\right\}_{j=1}^M
$$

\textbf{Loss Function}: We define a scalar-valued loss function \(L()\) to measure the prediction error. The mean square error (MSE) is used as the loss function in this work:
    \[
    \min _{f} \sum_{i=1}^L \sum_{j=1}^M \ell\left(\hat{F}_i^j, F_i^j\right)
    \]
    where \(L\) is the total number of time steps for each trajectory of measured state \(F\) and input \(x\), and \(\ell\) is the mean square error. Our loss function incorporates the output over the ODE integration time, ensuring that the model's predictions are consistent with the underlying physical reality.
    
\textbf{Backpropagation}: The Neural ODEs defines an adjoint state \(\mathbf{a(t)} =  \frac{\partial L}{\partial \mathbf{x_{t}}} \), and its derivation could be defined as:
    \[
    \frac{d \mathbf{a}(t)}{d t}=-\mathbf{a}(t)^{\top} \frac{\partial f(\mathbf{x}(t), t, \theta)}{\partial \mathbf{x}}
    \]
    The neural network \(f\) can use backpropagation to update its parameters.
    
\textbf{Optimization}: An optimization algorithm (e.g., Adam optimizer) is used to minimize the loss function and update the neural network parameters.

\section{Experiments}

\begin{figure}[htp]
    \centering
    \includegraphics[width=\linewidth]{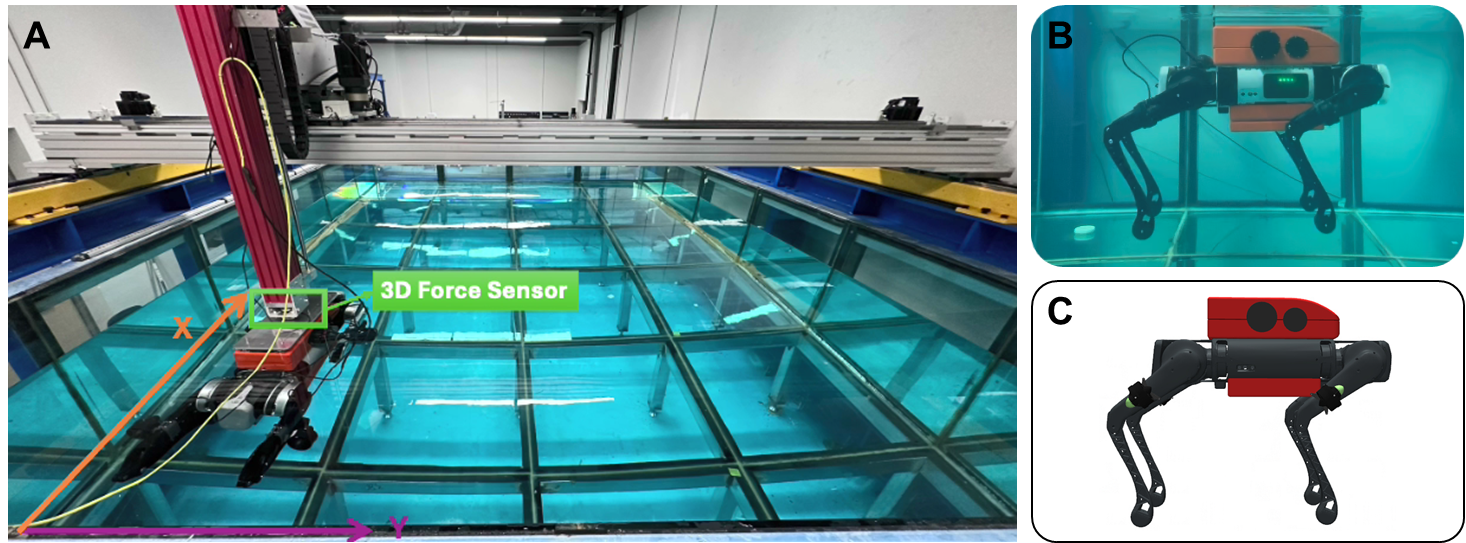}
    \vspace{-0.5cm}
    \caption{Experiment Setup. A: Experiment Setup. The 3D force sensor is attached between the gantry and robot. B: The side view of real robot in water tank. C: The side view of robot in simulation.}
    \label{fig:exp_setup}
    \vspace{-0.25cm}
\end{figure}

Training involves using a portion of the data to learn the model parameters, with separate validation and test sets used to evaluate generalization performance. 

\subsection{Setup}
\label{sec:setup}
To train and evaluate our model, we first collected a comprehensive dataset through towing experiments carried out in a controlled pool environment. The experimental setup is designed to measure the forces acting on the quadruped robot under various conditions. The setup includes the following components:

\textbf{Pool Environment}: A controlled pool environment ensures consistent and repeatable conditions for all experiments.

\textbf{Towing Mechanism}: A towing mechanism with adjustable speed settings is used to tow the robot at different speeds and directions.

\textbf{Force Sensors}: High-precision force sensors are attached to the robot to measure the forces along the x, y, and z axes.

\textbf{Robot Configuration}: The quadruped robot is configured with various limb joint angles to simulate different motion scenarios.

The experiment involves towing the robot at speeds ranging from 0.2 m/s to 0.5 m/s, with increments of 0.1 m/s, in three directions: x, y, and xy (45 degrees). The forces acting on the robot are recorded for each towing condition, providing a rich dataset for training and evaluating the hydrodynamic model. 
Fig.~\ref{fig:exp_setup} shows the experiment setup to collect hydrodynamic data. 

\subsection{Dataset}

\begin{figure}[htp]
    \centering
    \includegraphics[width=\linewidth]{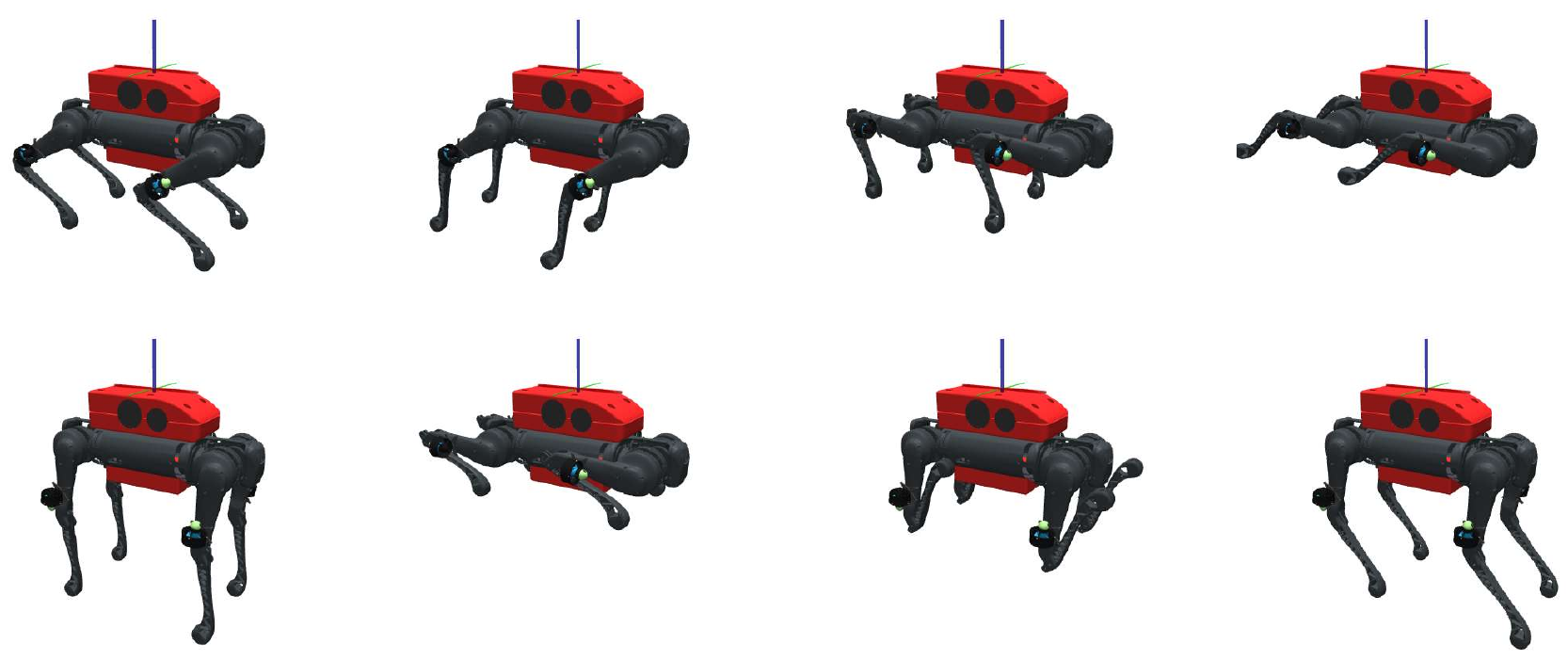}
    \vspace{-0.5cm}
    \caption{Robot configuration with different legs}
    \label{fig:leg_config}
    \vspace{-0.25cm}
\end{figure}

The dataset was collected during the towing experiments in Section~\ref{sec:setup}. The experiments were designed to measure the forces acting on the robot across 192 different towing speeds and the configuration of joints in Fig.~\ref{fig:leg_config}. 

To ensure that the simulation environment in MuJoCo accurately reflects the real-world performance of the robot, we implemented a multi-parameter optimization process to align the simulation with real data.

\textbf{Details of the Dataset} We use Unitree B1 quadruped robot \cite{unitree_b1_2024} and the joint limitations are based on official models. 
Due to the limitation of towing tank, the speed only has three conditions: X, Y, XY (45deg).
The input of the dataset is 4 dimensions, including joint position of second and third joint, linear speed in X and Y axes. The output is the hydrodynamic force in X and Y. We omit the force in Z that can be calculated based on gravity and buoyancy. 

\begin{figure}[htp]
    \centering
    \includegraphics[width=1.0\linewidth]{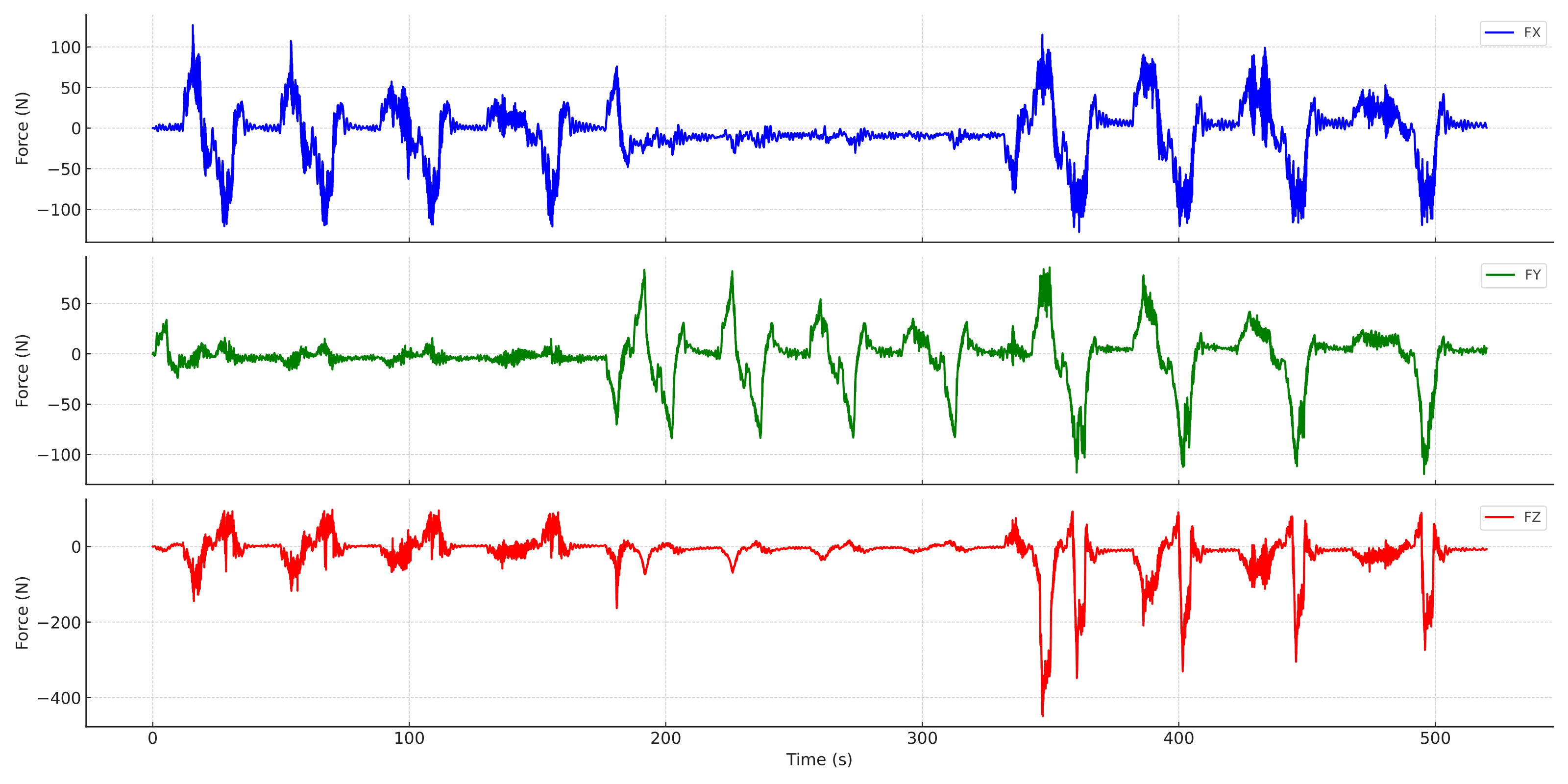}
    \caption{Raw data collected}
    \label{fig:raw_data}
\end{figure}

\textbf{Raw Dataset} Fig.~\ref{fig:raw_data} show the raw data collected from the force sensor during towing experiments. From the results we can see, there are three phrage, towing in axis X, Y and XY. Here we ignore data from Z-axis.

\textbf{Augmented Data} To enhance the dataset, we extend to more dimensions based on the real data collected, as shown in Table~\ref{table:dataset_aug}. From Table~\ref{Table:Datsets_setting}, our dataset is augmented as follows: \textit{Task 1.1} details the extension of the time series to 100-time steps, representing the sequential data of a quadruped robots maintaining a constant attitude angle. The Neural ODEs model is required to output the forces in two-axis directions throughout these 100-time steps. In \textit{Task 1.2}, to highlight the comparative predictive capabilities of the Neural ODEs under variable conditions for online learning, the temporal length of each condition was expanded to 10 time steps by randomly choosing from datasets. The condition variable was also concatenated, which varied across 5 distinct scenarios. \textit{Task 1.3}, addresses the increased complexity of conditions,  during this extension, random perturbations are introduced at each time step, with magnitudes equal to 10$\%$ of the standard deviation of the respective Force values. In \textit{Task 1.2}, and \textit{Task 1.3}, we resample the trajectory across 192 distinct conditions to investigate the effects of these switching conditions further. \textit{Task 1.2} is a challenging experiment involving 40 condition switches over 400 time steps, with a perturbation intensity of 10\%. 

\begin{table}[htbp]
    \centering
    \begin{tabular}{ccc}
        \toprule
        \textbf{Items} & \textbf{Unit} & \textbf{Dimension} \\
        \midrule
        Joint positions  & rad & 12 \\
        Joint velocities  & rad/s & 12 \\
        Quaternion  & - & 4 \\
        Angular velocity  & rad/s & 3 \\
        Linear velocity  & m/s & 3 \\
        Density  & kg/m³ & 1 \\
        \midrule
        Force in XYZ  & N & 3 \\
        Torque in XYZ  & Nm & 3 \\
        \bottomrule
    \end{tabular}
    \caption{Details of the Augmented Dataset}
    \vspace{-0.2cm}
    \label{table:dataset_aug}
\end{table}

\begin{table}[htp]
    \centering
    \begin{tabular}{c p{4.5cm} c}
        \toprule
        \textbf{Expr.} & \textbf{Input} & \textbf{Output}  \\
        \midrule
        \textit{Task 1.1} & Condition $\mathbf{x}$:[batch,$100$,4],  Initial: $F_{0}$[batch,2] & [batch,$100$,2] \\ 
        \midrule
        \textit{Task 1.2} & Condition $\mathbf{x}$:[batch,$50$,4],  Initial: $F_{0}$[batch,2] & [batch,$50$,2]  \\
        \midrule
        \textit{Task 1.3}  & Condition $\mathbf{x}$:[batch,$50$,4],  Initial: $F_{0}$[batch,2]& [batch,$50$,2] \\ 
        \midrule
        \textit{Task 2} & Condition $\mathbf{x}$:[batch,$400$,35],  Initial: $F_{0}$[batch,6]& [batch,$400$,6]\\ 
        \bottomrule
    \end{tabular}
    \begin{minipage}{\linewidth}
        \centering
        \vspace{10pt} 
        \footnotesize
        \vspace{-0.25cm}
        \textbf{Note:} batch refers to the batch size during the training stage
    \end{minipage}
    \vspace{-0.2cm}
    \caption{Input and output formats of the Datasets}
    \label{Table:Datsets_setting}
    \vspace{-0.5cm}
\end{table}

\subsection{Setting of Bechmark models}

To fairly compare the results of different baselines, we maintained consistent parameter counts as much as possible. We conducted comparative experiments on a custom dataset with attention-based, MLP-based, and LSTM-based models, as well as the CNODE model suggested by a reviewer(1Nsi). The motivation behind our experiment is twofold: 
\begin{itemize}
  \item To validate the advantages of the attention module by comparing it with the MLP-based ODE model. Is the inclusion of attention beneficial compared to not incorporating it?
  \item Highlight the strengths of our model by contrasting it with the LSTM and CNODE baseline models.  Is our model better than the baseline?
\end{itemize}

 \begin{table}[h]
    \centering
    \vspace{-0.25cm}
    
    \begin{tabular}[\linewidth]{p{1.5cm}p{1.35cm}p{1.35cm}p{1.35cm}p{0.85cm}} 
        \toprule
        \textbf{Models} & \textbf{MLP-ODE}  & \textbf{Attention-ODE} & \textbf{CNODE} & \textbf{LSTM}\\
        \midrule
        Hidden layer & [512,512,512]   & [512,512,512]  & [256,512,512]  &2\\
        Attention head  &None & 4   & None   &None \\
        Spline interpolation   &None & None  & Cubic& None\\
        Hidden state & None & None  & None & 256\\
        \bottomrule
    \end{tabular}
    \caption{Model settings}
    \label{tab:results}
    \vspace{-0.5cm}
\end{table}


\subsection{Model prediction performance }
\label{Model:E1}

In the following experiments, we record the performance of the Root Mean Squared Error (RMSE) and Mean Absolute Error (MAE) by predicting dynamic trajectories and ground truth. We compare standard models in the Neural ODEs framework, which include the Euler and RK4 integral method for ode and the Attention-based models proposed in this \textit{Task 1}.  

\begin{table*}[t]
    \vspace{-0.75cm}
    \caption{Performance on the different conditions within \textit{Task 1}}
    \centering 
    \label{Table:1}
    \begin{tabular}{ccccccc} 
        \toprule
        Models  & MAE-S  & RMSE-S   & MAE-C  & RMSE-C & MAE-N  & RMSE-N  \\ 
        \midrule
        MLP-ODE-euler & 9.2e-3 & 3.8e-3 &3.3e-3 & 4.7e-3 & 4.9& 6.4\\ 

        \midrule
        Attention-ODE-euler & 8.1e-3 & 4.1e-3 & 3.1e-3& 2.1e-3& 3.7&5.0\\ 
        \midrule
        MLP-ODE-RK4  & 4.9e-4  & 6.0e-4 &9.2e-4& 4.6e-4& 2.6& 4.7\\ 
        \midrule 
        Attention-ODE-RK4 & \textbf{2.7e-4*} & \textbf{6.0e-4*} &\textbf{3.3e-4*} & \textbf{6.4e-5*} & \textbf{2.1*}&\textbf{4.2*}\\ 
        \bottomrule
    \end{tabular}
    \begin{minipage}{\textwidth}
        \centering
        \vspace{5pt} 
        \footnotesize
        \textbf{Note:} The suffix '-S' indicates static conditions in \textit{Task 1.1}, '-C' denotes conditions that change over time in \textit{Task 1.2}, and '-N' represents noisy and changing conditions in \textit{Task 1.3}. '*' means the best performance.
    \end{minipage}
    \vspace{-0.5cm}
\end{table*}

From Table~\ref{Table:1}, it is evident that for ODE transition problems, attention-based methods generally outperform those without attention. This improvement is likely due to the integration of localized attention mechanisms focusing more on conditions. Additionally, our attention methods perform well under noisy conditions, with errors of 4.2, suggesting significant potential for model deployment. From an integration perspective, there is minimal difference between the euler and RK4 methods in noise-free scenarios. However, in the noisy setting of Expr3, the RK4 method exhibits minor  errors. 

In \textit{Task 2}, we focused on evaluating the performance of several baseline models, specifically the MLP-ODE, CNODE, and LSTM models. The prediction results for the output are presented in Table~\ref{tab:baselineresult}.

\begin{table}[htp]
    \centering
    \caption{Performance of different models in the \textit{Task 2}}
    \label{tab:baselineresult}
    \begin{tabular}{ccccc} 
        \toprule
        Metric &MLP-ODE  &\underline{Attention-ODE}  &CNODE &LSTM\\
        \midrule
        Time  & 17ms   & \underline{15ms}  &26ms  &80ms\\
        Parameters  & 530628   & 554934   &636626 & 828422 \\
        MAE   & 29.4   & 18.6  & 19.7 & \underline{16.3} \\
        RMSE   & 43.1  & 40.2  & \underline{38.0} & 43.9\\
        \bottomrule
    \end{tabular}
    \vspace{-0.25cm}
\end{table}

\begin{figure}[htp]
    \centering
    \includegraphics[width=\linewidth]{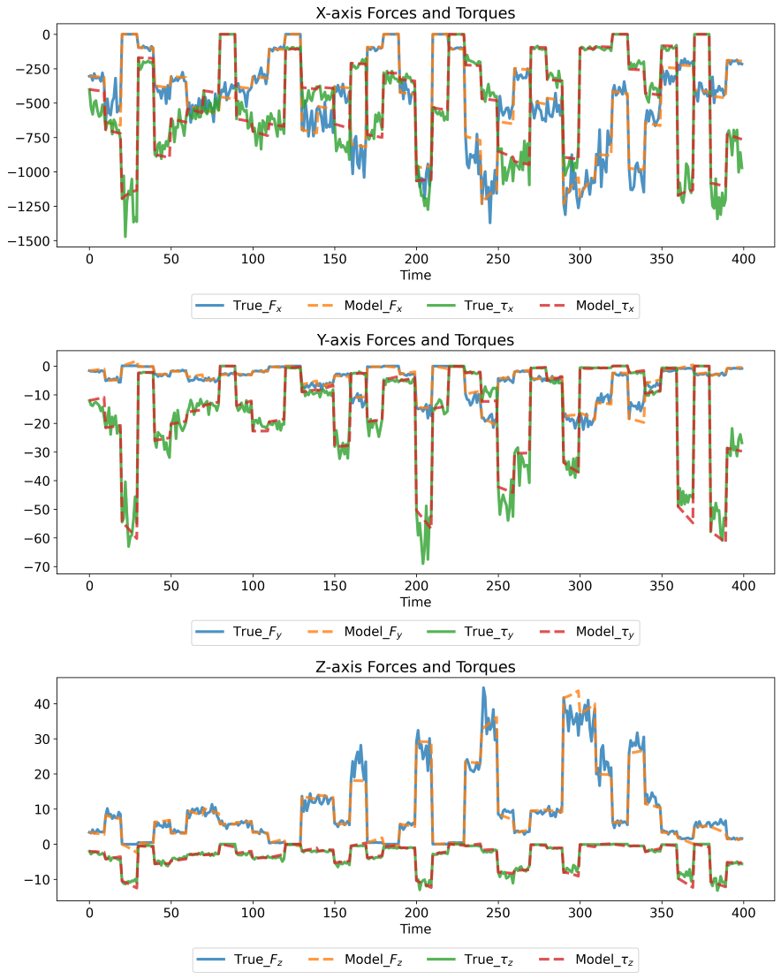}
    \caption{Comparison of the baseline methods in \textit{Task 2}}
    \label{fig:final}
    \vspace{-0.35cm}
\end{figure}

 Comparative experiment results as shown in the Table~\ref{tab:baselineresult} and Fig.~\ref{fig:final}. Firstly, MLP-based and attention-based ODEs can handle long-time sequences more efficiently, with inference times of 17ms and 15ms, respectively. The results suggest that the slower speed of CNODE is due to the need for spline interpolation of the data. While this operation is beneficial for irregular time sequences, it becomes a time-consuming step for regular time sequences. For consistency, we fixed the integration method for all three ODEs using the forward Euler method. Secondly, compared to traditional methods like LSTM. Although the LSTM model has a relatively lower MAE, the result shows that its predictions exhibit more fluctuations, indicating that it has learned more noise rather than the physical law. The attention-based model exhibits shorter inference times. Considering inference speed, model parameters, and metrics such as MAE and RMSE, attention-based methods emerge as a promising choice.

To verify the model further, we test it in the MuJoCo simulator, as shown in Fig.~\ref{fig:swimming_mujoco}. More details can be found in the supplementary videos.

\begin{figure}[h]
    \centering
    \includegraphics[width=0.9\linewidth]{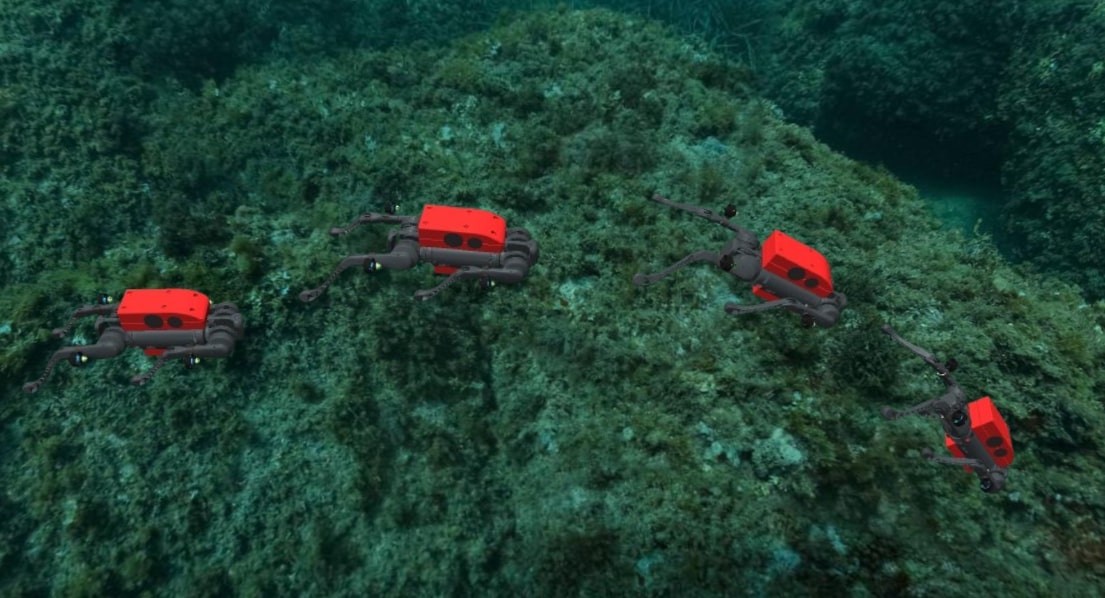}
    \caption{Test model in simulation}
    \label{fig:swimming_mujoco}
    \vspace{-0.75cm}
\end{figure}

\section{Conclusion}
        
This paper proposes a data-driven model to learn the complex hydrodynamic model. We have developed attention-based Neural ODEs for dynamic prediction in underwater quadruped robots. Using data augmentation, our model uses kinematic trajectories as input and outputs dynamic hydrodynamic forces.
The proposed model not only reduces computational overhead compared to traditional methods but also enhances the robot's autonomous behavior in dynamic environments. This work contributes significantly to the field of underwater robotics by providing a scalable and efficient solution for real-time applications.

\textbf{Future Work} The model's reliance on initial value calibration may necessitate prior estimates of fluid resistance coefficients in real-world scenarios. Further tests will be conducted in real-world environments to validate the model's performance.
Testing on real robots will be conducted in the future.






\clearpage
\newpage

\bibliographystyle{unsrt}
\bibliography{reference}

\end{document}